 \documentclass{article}

     \PassOptionsToPackage{numbers, compress}{natbib}
 
 \usepackage[final]{neurips_2023}

 \usepackage[utf8]{inputenc} 
 \usepackage[T1]{fontenc}    
 \usepackage{hyperref}       
 \usepackage{url}            

 \usepackage{booktabs}       
 \usepackage{amsfonts}       
 \usepackage{nicefrac}       
 \usepackage{microtype}      
 \usepackage{xcolor}         
 \usepackage{graphicx}
 \usepackage{svg}
\usepackage{amsmath,amsfonts,bm}
 
 \author{
 Gül Sena Altıntaş \hspace{2em} 
 Gregor Bachmann \hspace{2em} Lorenzo Noci \hspace{2em} Thomas Hofmann\\
 ETH Zürich\\
 \texttt{\{galtintas, gregorb, lnoci\}@ethz.ch}
 }
 
\usepackage{amsmath,amsfonts,bm}

\def\eqref#1{equation~\ref{#1}}

\def\1{\bm{1}}

\DeclareMathAlphabet{\mathsfit}{\encodingdefault}{\sfdefault}{m}{sl}
\SetMathAlphabet{\mathsfit}{bold}{\encodingdefault}{\sfdefault}{bx}{n}

 \usepackage{mathtools}
 \usepackage{booktabs}
 \usepackage{subfigure}  
 \usepackage{subcaption}
 \usepackage{caption}
 \usepackage{multirow}
 \usepackage{multicol}
 \usepackage{wrapfig} 
 \usepackage{tikz}
 \newcommand{\acc}[2][]{\ifmmode \text{Acc}_{#1}(#2) \else Acc$_{#1}$(#2) \fi}

 \title{Disentangling Linear Mode Connectivity}

\begin{document}
 
 \newcommand{\sectionSkip}{\vspace{70pt}}
 \maketitle
 
 \begin{abstract}
     Linear mode-connectivity (LMC) (or lack thereof) is one of the intriguing characteristics of neural network loss landscapes. While empirically well established, it unfortunately still lacks a proper theoretical understanding. Even worse, although empirical data points are abound, a systematic study of when networks exhibit LMC is largely missing in the literature. In this work we aim to close this gap. We explore how LMC is affected by three factors: (1) architecture (sparsity, weight-sharing), (2) training strategy (optimization setup) as well as (3) the underlying dataset. We place particular emphasis on minimal but non-trivial settings, removing as much unnecessary complexity as possible. We believe that our insights can guide future theoretical works on uncovering the inner workings of LMC.
 \end{abstract}

 \section{Introduction}
          
 In recent years, there has been a growing interest in understanding the geometry of loss landscapes, how modern stochastic first-order gradient-based algorithms navigate them and the relationship between different optima. There is a large body of work on the mode-connectivity (MC) \cite{garipovLossSurfacesMode2018, draxlerEssentiallyNoBarriers2018,bentonLossSurfaceSimplexes2021}, \textit{linear} mode-connectivity (LMC) \cite{frankleLinearModeConnectivity2020}, permutation invariance \cite{entezariRolePermutationInvariance2022, ainsworthGitReBasinMerging2023, benzingRandominitialisationsPerforming2022, simsekGeometryLossLandscape2021} and a broader range of symmetries \cite{, zhaoSymmetriesFlatMinima2023} of neural networks, showing that loss landscapes are not solely characterized by high non-convexity and isolated minima but can often contain flat connected regions. A more detailed account of these works are included in \autoref{ap:sec:related-work}.
 
Crucial to this work, LMC (and the lack of it) has been observed in disparate settings, however its root causes have not been epistemically investigated. For instance, when it comes to architectural components, it is well-known that convolution-based architectures lack LMC \cite{frankleLinearModeConnectivity2020} compared to fully-connected models even after accounting for permutation invariance \cite{entezariRolePermutationInvariance2022, ainsworthGitReBasinMerging2023}. The varying factors distinguishing these two architectures, locality, weight-sharing, pooling layers, etc., make it hard to pinpoint the source of disruption for LMC. Furthermore, these architectures are often trained under different optimization schemes and with different datasets, which are likely to be confounding factors.
 
 In this paper, we systemically isolate some of the causes of LMC. We start from the simplest connected setting of logistic regression, i.e. linear model with no hidden layers, and gradually incorporate architectural changes, training techniques, and datasets typically used in modern deep learning pipelines. We identify the minimal non-linear setting, namely an MLP with one hidden layer and ReLU activation, where LMC can be robustly observed over different optimization schemes and datasets. We then analyze which components break LMC, in particular we study:
 \begin{itemize}
     \item The effect of the model architecture by introducing locality, weight-sharing, and sparsity to the hidden layer. This way, we recover locally connected, convolutional, and attention-based models that have a correspondence with the minimal model. Our experiments suggest that while locality preserves LMC, weight-sharing breaks it. 
     \item Optimization algorithm and training strategy.
     We show that ADAM breaks connectivity more than SGD, while it can be recovered by modifying learning rate and adding warm-up. 
     \item How dataset complexity affects LMC 
     by training MLPs with increasing dataset complexity, namely MNIST, CiFAR-10, CiFAR-100 and TinyImageNet. We observe that LMC can be more easily broken under more complex datasets.
 \end{itemize}
 
 \section{Background}
 We consider the classification problem for a general $L$-layer model with $\sigma$ activation trained with the cross entropy loss, whose intermediate output at layer $l < L$ is given by:
 $$\mathbf{z}_l \coloneqq f_l(\mathbf{x}; \mathbf{\theta}_l) = \sigma(W_l \mathbf{z}_{l-1} + \mathbf{b}_l)$$
 
 and the final output is $\hat y \coloneqq W_L \mathbf{z}_{L-1} + \mathbf{b}_{L}$. When $L=1$ we recover logistic regression, and if $\sigma$ is the identity function we have an $L$-layer linear model. All models in the rest of the text are trained for 200 epochs and non-linear models reach $\sim 0$ training loss.
 
 \paragraph{Linear Interpolation:} For two networks $A$ and $B$ with parameters $\Theta_A$ and $\Theta_B$, their linear interpolation is defined with respect to the convex combination of the parameters at each layer, i.e. $\Theta(\alpha) \coloneqq \{(1-\alpha) W_{A_i} + \alpha W_{B_i}, (1-\alpha) \mathbf{b}_{A_i} + \alpha \mathbf{b}_{B_i} \}_{i: 1\to L}$.
 
 \paragraph{Error Barrier:}
 We are interested in how the error evolves along the linear path between two models $A,B$, where $\Theta_B \coloneqq \Theta_A$, during training. Do they stay linearly mode-connected even though they are trained separately, i.e. with different SGD noise (data orderings and augmentation). 
 We base our measure of connectivity on 
 \cite{frankleLinearModeConnectivity2020}'s definition of the error barrier (\autoref{background:eq:barrier}). 
 
 \begin{align}
     \mathcal{B} &= \sup_\alpha \mathcal E(f(\cdot; \Theta(\alpha))) - \frac 1 2 (\mathcal E(f(\cdot; \Theta_A)) + \mathcal E(f(\cdot; \Theta_B)))
     \label{background:eq:barrier} 
 \end{align}
 The error is quantified as the ratio of incorrect predictions, represented as $\mathcal E(\cdot) \coloneqq (1-\acc{\cdot}), \; \acc{\cdot} \in [0, 1]$.
 While the current barrier definition offers an absolute measure, it doesn't differentiate the extent of performance loss, which is the primary focus of LMC research, across various levels of task complexity.  Hence, we propose to use a normalized version that accounts for test accuracy when comparing the same architecture on different datasets.
 \begin{equation}
     \Bar{\mathcal{B}} = \frac{\mathcal B}{\frac 1 2 (\acc[te]{f(\cdot; \Theta_A)} + \acc[te]{f(\cdot; \Theta_B)})}
     \label{background:eq:performance-barrier}
 \end{equation}
 
 We follow the convention in the literature and evaluate $\Theta(\alpha)$ at $T$ equidistant values of $\alpha$ between 0 and 1. 
 T is set to 11, i.e. the model is evaluated at $\alpha \in \{0, 0.1, \dots, 1\}$.  We refer to a model as linearly mode-connected if two independent runs of SGD starting from the \textit{same} random initialization exhibit low barrier, e.g. $< (0.02)$.

 \begin{table}[h]
 \centering
 \caption{Summary of training and testing error barriers in percentage $(100\cdot \mathcal{B})$ for an $L$-layer linear model across various optimization schemes (optimizer/learning rate/batch size(BS)). Each model reaches train cross entropy between 0.2 and 0.3 and above 90\% test accuracy except for the high learning rate regime noted with $\ast$, where the training is unstable.}
 \label{exp:tab:linear-barriers}
 \begin{tabular}{lrlrlrlrl}
 \toprule
                         & \multicolumn{4}{c}{SGD}           & \multicolumn{4}{c}{ADAM}          \\
  & \multicolumn{2}{c}{High Lr (0.1)} & \multicolumn{2}{c}{Med Lr (0.01)} & \multicolumn{2}{c}{High Lr (0.005)} & \multicolumn{2}{c}{Med Lr (0.001)} \\
                         & Test   & Train  & Test   & Train  & Test   & Train  & Test   & Train  \\\midrule
 1-Layer, BS=30K  & 0.05 & 0.01 & 0.00 & 0.00 & 0.03 & 0.00 & 0.01 & 0.01 \\
 1-Layer, BS=1024        & 0.06 & 0.04 & 0.06 & 0.00 & 0.12 & 0.07 & 0.08 & 0.05 \\
 
 2-Layer, BS=30K  & 0.01 & 0.01 & 0.00 & 0.00 & 0.11 & 0.14 & 0.05 & 0.01 \\
 2-Layer, BS=1024        & 0.01 & 0.14 & 0.05 & 0.09 & 1.30 & 1.45 & 0.18 & 0.16 \\
 4-Layer, BS=1024        & 0.06 & 0.06 & 0.02 & 0.01 & 47.74 & 48.91 & 0.07 & 0.14 \\
 8-Layer, BS=1024        & $\ast$      & $\ast$      & 0.06 & 0.00 & \textbf{68.22} & \textbf{68.87} & \textbf{73.78} & \textbf{74.32} \\
 \bottomrule
 \end{tabular}
 \end{table}
 
 \section{Finding a minimal model}
 \paragraph{Logistic regression:} The simplest model is given by logistic regression, i.e. a linear model with no hidden layers. We are interested in studying LMC as the complexity of the underlying setting increases, thus logistic regression is an intuitive starting point.
 Since it is a convex problem, we expect it to satisfy LMC (even for \textit{different} initializations).
 We show the results in the first row of \autoref{exp:tab:linear-barriers}.
 We indeed confirm this empirically, as the model remains connected for all datasets, batch sizes and optimizers.
 
 \paragraph{Linear networks:} 
 Next, we study how the dynamics change when more layers are added while keeping the network linear. We show the analogous results for several depths $L \in \{2, 4, 8\}$ in \autoref{exp:tab:linear-barriers}. Surprisingly, we observe that for SGD with momentum, linear networks remain very connected even up to $8$ layers. ADAM on the other hand quickly breaks connectivity even for shallower networks. We attribute the difference to ADAM's adaptivity. This already hints at a re-occurring theme in this work; for all the considered settings, ADAM tends to amplify the resulting barriers.
 
 \paragraph{Non-linear networks:} 
 
\begin{figure}[!htbp]
     \centering
     \foreach \depth in {2,4,8}{
         \begin{subfigure}{}
             \includegraphics[width=0.29\textwidth]{./sgd-\depth-layer-mlp-leaky-slopes.pdf}
         \end{subfigure}
     }
     \begin{subfigure}{}
         \includegraphics[width=0.7\textwidth]{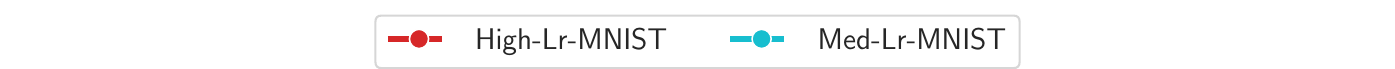}
     \end{subfigure}
     \caption{Barrier with respect to the negative slope $p$ of the Leaky ReLU for $p\in \{0, 0.01, 0.1, 0.2, \dots 0.9, 0.99\}$. It is equivalent to ReLU for $p=0$ and Identity for $p=1$.}    \label{exp:fig:leaky-relu}
 \end{figure}

 We now consider the role of the non-linearity. Since ADAM already breaks connectivity even for the linear setting, we focus on SGD in order to avoid confounding.  We gradually turn the network non-linear by taking Leaky-ReLU with various slopes ranging from $1$ (linear) to $0$ (ReLU). We display connectivity as a function of the slope in \autoref{exp:fig:leaky-relu} for MLPs of different depths. Non-linearity coupled with larger depth enlargens the barriers, even for SGD with momentum, highlighting that non-linearity has a detrimental effect on connectivity. The 1 hidden layer case however remains surprisingly robust in terms of connectivity. 
 
 \paragraph{Minimal model:} Given its strong connectivity values and non-trivial nature, we will adopt the 1 hidden-layer ReLU MLP trained with SGD as our \textbf{minimal} model which still exhibits LMC. We will show in the following, how various interventions such as optimizer choice, architectural changes as well the dataset can affect LMC. This gives us a clean minimal setting to disentangle different effects.
 
 \section{Interventions to minimal model}
 \paragraph{Training Strategy:}
 
 \autoref{exp:tab:linear-barriers} already suggests that the optimization algorithm plays an important role in connectivity. We attempt to decouple its effect further over three dimensions: (1) Choice of the algorithm: SGD or ADAM (2), Learning Rate: High (0.005 for ADAM and 0.1 for SGD) denoted by $\uparrow$, and Medium (0.001 for ADAM and 0.01 for SGD), denoted by $\downarrow$ and (3) Warmup: either no warmup or linear warmup for 10 epochs, i.e. 5\% of total training time. We display the resulting connectivity values when varying the minimal model along the outlined factors in \autoref{tab:connectivity-opt}. We again observe a very similar pattern; Switching to the ADAM optimizer results in a significantly less connected model. Using warm-up on the other hand leads to a significant reduction in barrier, suggesting that LMC is determined early in the training. 
 
 \begin{table}[]
     \centering
     \caption{Error barriers presented in percentage when changing the optimization setup for the minimal model, MLP with one hidden layer and ReLU activation.}
     \label{tab:connectivity-opt}
     \begin{tabular}{lccccc}
     \toprule
      & SGD $\uparrow$ Lr & SGD $\uparrow$ Lr W-up & ADAM $\downarrow$ Lr & ADAM $\uparrow$ Lr & ADAM $\uparrow$ Lr + W-up \\
      \midrule
     $\% \mathcal{B}_{\text{train}}$   & 0.00&0.00&0.00&2.92&0.17  \\
     $\% \mathcal{B}_{\text{test}}$   & 0.05&0.02&0.06&2.86&0.23  \\[2mm]
     \bottomrule
     \end{tabular}
 \end{table}

 \paragraph{Architecture:}

 We now explore the role of the layer-type in the $1$-hidden layer model. We are especially interested in how \textit{locality} and \textit{weight-sharing} affect LMC in this simple model. Let us denote the underlying weight matrix by $W \in \mathbb{R}^{m \times d}$ where $d$ is the input dimension and $m$ the number of hidden units. 
 
 \begin{wrapfigure}{r}{0.33\textwidth}
     \begin{subfigure}
         \centering
          \includegraphics[width=0.28\textwidth]{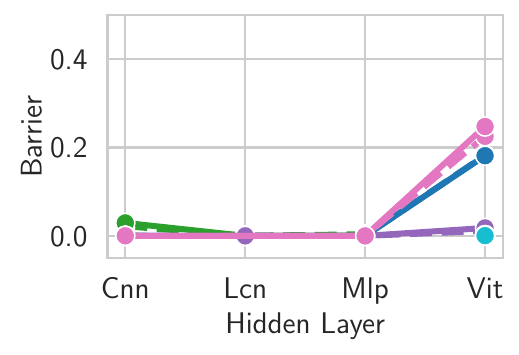}
     \end{subfigure}
     \begin{subfigure}
         \centering
         \includegraphics[width=0.3\textwidth]{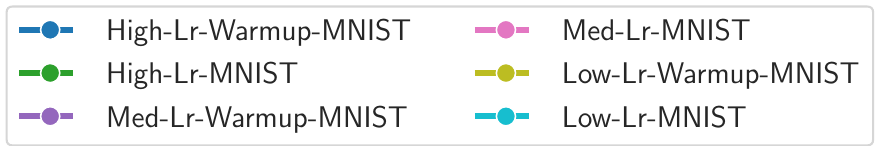}
     \end{subfigure}
     \caption{Error barriers when varying the hidden layer. Note that previous SGD learning rates were too high for the ViT, hence we introduce the \textit{Low-Lr} of 0.001 where ViTs are also connected}
     \label{exp:fig:architecture}
 \end{wrapfigure}
 It is simple to show  that a locally-connected CNN (exhibiting locality) and a CNN (exhibiting both locality and weight-sharing) can be obtained by imposing sparsity on ${W}$ and tying its parameters correctly (see \autoref{ap:sec:arch} for details). Using this correspondence between MLP, LC-CNN and CNN we can understand the effect of these structural choices in a controlled manner. For completeness, we also experiment with using a single attention layer. While in this case we do not have a direct
 correspondence, attention still offers both locality and weight-sharing. We show the effects of such architectural changes in \autoref{exp:fig:architecture}. We observe that the LC-CNN remains very connected but both the CNN and attention-based model experience a decay in connectivity. These results suggest that weight-sharing might play a more important role for connectivity than previously appreciated. 
 
 \paragraph{Role of dataset:}
 
\begin{figure}[!htbp]
     \centering
     \foreach \depth in {2,4,8}{
         \begin{subfigure}
             \centering
             \includegraphics[width=0.29\textwidth]{./datasets-sgd-\depth-layer.pdf}
         \end{subfigure}
     }
     \hfill
     \begin{subfigure}
     \centering
         \includegraphics[width=0.7\textwidth]{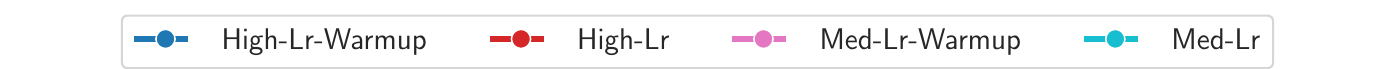}
     \end{subfigure}
     \caption{{Effect of dataset complexity on LMC} on an $L$-layer MLP. Note that to account for the task complexity we compare LMC using the normalized performance-aware barrier (\autoref{background:eq:performance-barrier}). 
     }
     \label{fig:exp:dataset}
 \end{figure}

 So far, we focused on the MNIST dataset. We now investigate how increasing task complexity affects LMC for our minimal model. To account for the performance gap we use the normalized performance-aware barrier (\autoref{background:eq:performance-barrier}). We elaborate on the different task complexities in \autoref{ap:sec:data}. In \autoref{fig:exp:dataset}, we observe that task complexity hinders LMC when accounted for the overall performance. 
 The more complex the dataset, the larger the barrier and the difference between the training and test barriers.

 \section{Discussion} 
 In this work, we examined how each individual component, training strategy, architecture and dataset impact LMC of two networks that are trained from the same initialization with different SGD noise. We identified a minimal but non-trivial model amenable to theoretical analysis which show-cases precisely how several factors such as (1) optimization setup, (2) architectural design ,and (3) dataset choice influence connectivity. We believe that our results can serve as a guide for theoretical progress in this topic, equipping the theorist with a model that is very simple but at the same time very rich in phenomenology. We thus hope that future work can build upon our empirical findings.

 \bibliography{references}
 \bibliographystyle{plainnat}
 
 \clearpage
 \appendix
 \section*{Appendix}
 
 \section{Reproducibility Statement}
 We used the FFCV-SSL package by \cite{bordes2023ffcv_ssl} built on \cite{leclerc2023ffcv}'s FFCV package to ensure full reproducibility in terms of the SGD noise, see \autoref{ap:sec:data_reproduce} for more details.

 \subsection{Data Loader Reproducibility \label{ap:sec:data_reproduce}}
\begin{figure}[h]
     \centering
     \foreach \seed in {43,118}{
     \begin{subfigure}
         \centering\includegraphics[width=0.9\textwidth]{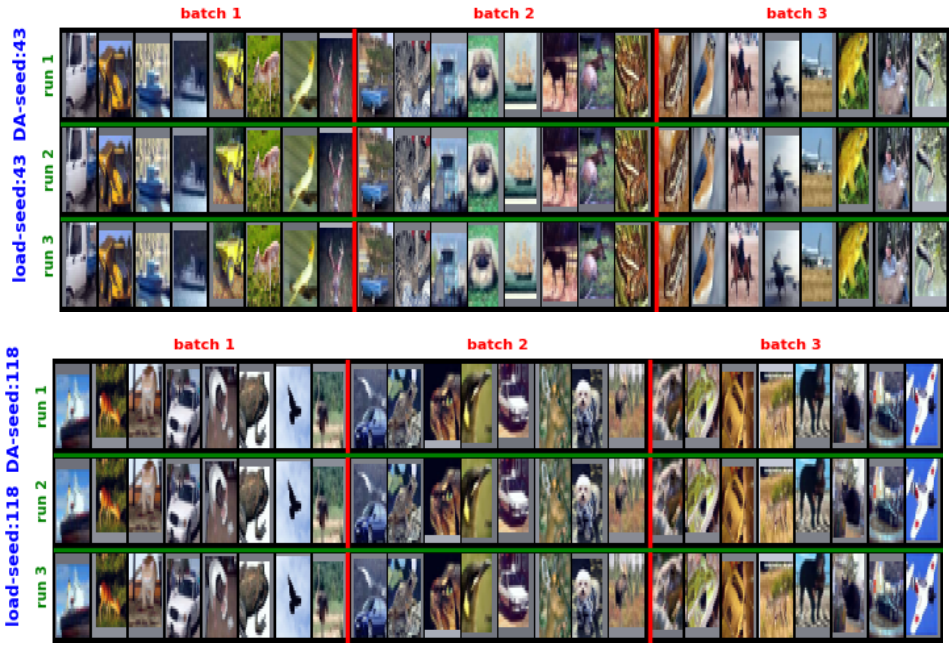}
     \end{subfigure}
     }
     \caption{Each data loader is initialized three times with loader seed (used for data ordering) and data augmentation seed (used for random augmentations) set to 43 (top) and 118 (bottom). Augmentations used: random translate and horizontal flip}
     \label{ap:fig:data_reproduce}
 \end{figure}

 The data ordering affects stochastic gradient based methods and hence connectivity. To ensure our runs can be reproduced we pick a trainer seed for initialization, data loader seed that determines ordering and an augmentation seed used for random augmentations. We include two sample CIFAR-10 data loaders in \autoref{ap:fig:data_reproduce}.
 
 \section{Related Work \label{ap:sec:related-work}}
 This appendix provides a brief overview of the relevant literature.

 \textbf{Mode-Connectivity:} \cite{garipovLossSurfacesMode2018, draxlerEssentiallyNoBarriers2018} demonstrated that optima trained from different initializations can be connected with simple parametric curves, e.g., polygonal chains or Bezier curves, without incurring a significant increase in the loss along this path. \cite{bentonLossSurfaceSimplexes2021} showed that these paths can be extended to probabilistic volumes of low loss. \cite{simsekGeometryLossLandscape2021}  formalized these volumes for over-parameterized two-layer networks as the \textit{Global Minima Manifold} and provided explicit descriptions of its dimensions.
\cite{vlaarWhatCanLinear2022} analyzed the optimization trajectory on the loss landscape by linearly interpolating between the initial and final training weights.
 
 \textbf{Linear Mode-Connectivity:} A parallel line of work, explore \textit{linear} mode-connectivity (LMC), where a linear path of near-constant error exists between the two optima. \cite{frankleLinearModeConnectivity2020} show that two fully-connected networks trained from the same initialization but with different SGD noise, i.e. data order and augmentations, are stable to the noise and converge to linearly connected minima. Their results extend to more complex vision algorithms as well if the two networks are trained with the same SGD noise for a while and then spawned.
 More recently \cite{zhouGoingLinearMode2023} coined Layerwise Linear Feature Connectivity (LLFC), a stronger setting where the feature maps of every layer also exhibit LMC. \cite{junejaLinearConnectivityReveals2023, neyshaburWhatBeingTransferred2021, lubanaMechanisticLensMode2023} study generalization in transfer learning through LMC.

 \textbf{Permutation Invariance}: 
 The permutation symmetry of the hidden neurons and how the learned parameters interact with each other after accounting for the permutation has also emerged as a notable avenue of inquiry. 
 \cite{entezariRolePermutationInvariance2022}'s initial conjecture argued that in most cases SGD converges to the same basin up-to permutation and showed the emergence of LMC for wide and shallow architectures after accounting for the permutation invariance. \cite{ainsworthGitReBasinMerging2023} proposed a general weight matching algorithm to align models trained from different initializations that supported \cite{entezariRolePermutationInvariance2022}'s conjecture on ResNets as well.
 \cite{benzingRandominitialisationsPerforming2022} show that two models exhibit linear mode-connectivity at initialization when merged with the permutation found later in training.
 
 \section{Further Training Strategies \label{ap:sec:training}}
 This appendix presents some ablation studies regarding the training techniques and optimization. 
 
 Although we examined the effect of other regularization techniques, our preliminary experiments proved these three dimensions, optimizer, learning rate, and warm-up, to be the most important for LMC across different architectures. For example, varying the batch size, turning off momentum, adding a weight decay term or cosine learning rate scheduler doesn't have a significant impact on the behavior of the previous settings.
 We found that gradient clipping can also be used to preserve LMC. 
 
 \subsection{ADAM on MLPs}
\begin{figure}[h]
     \centering
     \begin{subfigure}
         \centering
         \includegraphics[width=0.3\textwidth]{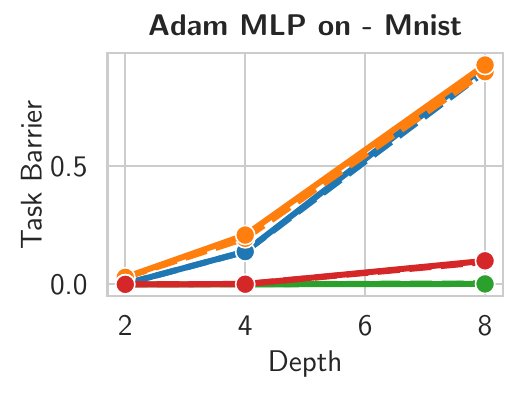}
     \end{subfigure}
     \begin{subfigure}
         \centering
         \includegraphics[width=0.3\textwidth]{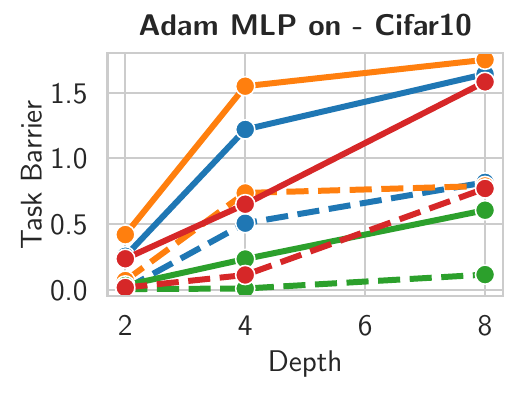}
     \end{subfigure}
     \hfill

     \begin{subfigure}
         \centering
         \includegraphics[width=0.75\textwidth]{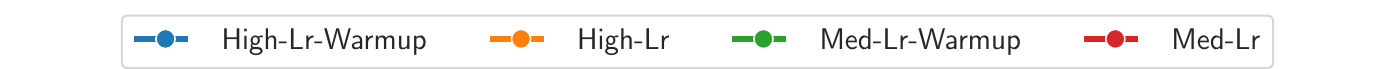}
     \end{subfigure}
     \caption{Task barrier for MLPs trained with ADAM on MNIST (left) and CiFAR-10 (right)}\label{ap:fig:adam-mlp-mnist-cifar10}
 \end{figure}

 Since ADAM already breaks LMC in deeper linear models, we find it trivial that it also doesn't preserve LMC in MLPs. Still, for completeness we provide the performance-aware barrier for MNIST and CiFAR-10 in \autoref{ap:fig:adam-mlp-mnist-cifar10}.
 
 \section{Architectural Correspondence\label{ap:sec:arch}}
 Similar to \cite{neyshaburLearningConvolutionsScratch2020}, we study shallow convolutions and establish their MLP counterpart based on the Toeplitz representation of the underlying convolutional layer. For simplicity, we set stride equal to the kernel size and do not use 0-padding. A convolutional layer operating on a $C_i \times H \times W$ input with kernel size $(k_1, k_2)$ and $C_o$ filters has $C_o \times C_i \times k_1 \times k_2$ parameters. Its locally connected counterpart uses different kernels to compute each target pixel, hence it has $C_o {\color{red} \times  H^\prime \times W^\prime}\times C_i \times k_1 \times k_2$ parameters, where ${\color{red} H^\prime, W^\prime}$ is the output spatial dimension of the resulting feature map. Both of them can be embedded in a $C_o \cdot H^\prime \cdot W^\prime \times C_i \cdot H \cdot W$ linear layer.
 \autoref{tab:param_count} shows the total number of parameters for a 2-Layer network, where the first layer is either a convolution, locally connected or linear layer whose weights can be represented with the same Toeplitz matrix.
 Note that ViTs \cite{dosovitskiyVIT2021} could also be considered in this framework thanks to \cite{cordonnierRELATIONSHIPSELFATTENTIONCONVOLUTIONAL2020}, however we leave it to future work to study the exact correspondence.
 
 \begin{table}[h]
     \centering
     \caption{Parameter Count (M) for a 2-Layer Network where the first layer is either a CNN/LC-CNN/Linear equivalent to kernel size=4, padding=0}
     \label{tab:param_count}
     \begin{tabular}{lccccc}
     \toprule
         CNN  & 0.09 \\
         LC-CNN & 0.48 \\
         MLP & 25.26 \\
     \bottomrule
     \end{tabular}
 \end{table}
 
 \paragraph{ViT-like MLP}
 To study the effect of attention on LMC, we consider the simplest setting of a ViT. We don't modify the patch embeddings, normalizations and the classifier layer but simplify the encoder. We remove skip connections and the MLP part (last two linear layers) from the transformer encoder block and only use one block. We use patch size of 4 to establish similarity to the (LC)-CNN case. 8 heads each of dimension 48.
 The resulting architecture has approximately $ 1.08$M parameters.
 
 \section{Data \label{ap:sec:data}}
 In \autoref{fig:exp:dataset}, we gradually increase the complexity of the task by changing the dataset: 
 \begin{enumerate}
     \item \textit{MNIST $\to$ CiFAR-10} input dimensions (both spatial and number of channels) increase from $(28, 28, 1)$ to $(32, 32, 3)$ while keeping the number of target labels the same. These two datasets also have similar number of samples (60,000 and 50,000).
     \item \textit{CiFAR-10 $\to$ CiFAR-100} number of samples and image resolution stay constant while the number of labels increase by a factor of 10, from 10 to 100.
     \item \textit{CiFAR-100 $\to$ Tiny-ImageNet} image resolution, number of labels and number of samples double.
 \end{enumerate}
 
 We limit this analysis to 2-4-8-Layer MLPs trained using SGD with momentum with high (0.1) or medium (0.01) learning rate. Since we are interested in the most simple settings, we don't use any data augmentation, which hurts generalization. Moreover, MLPs are known for their subpar performance on large scale image classification tasks. See \autoref{ap:tab:test_accuracies} for a comparison of the test accuracies across these four datasets. We propose \autoref{background:eq:performance-barrier} to account for this performance gap. This modification allows us to view error barrier as a ratio of the lost performance.
 
 \begin{table}[h]
 \centering
 \caption{Test accuracies (\%) reached on varying datasets by $L$-Layer MLPs trained using SGD with high (0.1) or medium (0.01)}
 \label{ap:tab:test_accuracies}
 \begin{tabular}{lrlrlrlll}
 \toprule
              & \multicolumn{2}{c}{2-Layer}   & \multicolumn{2}{c}{4-Layer}   & \multicolumn{2}{c}{8-Layer}   &  &  \\\midrule
              & High & Med & High & Med & High & Med &  &  \\
 MNIST        & 98.32        & 98.34        & 98.41        & 98.19        & 98.41        & 97.14        &  &  \\
 CiFAR-10     & *             & 54.24        & 58.75        & 55.51        & 57.44        & 54.45        &  &  \\
 CiFAR-100    & *             & 26.01        & 14.26        & 27.16        & 25.33        & 20.30        &  &  \\
 TinyImageNet & *             & 7.62        & 1.68        & 8.27        & 5.80        & 5.79        &  &  \\
 \bottomrule
 \end{tabular}
 \end{table}
 
 
 \end{document}